\documentclass[letterpaper,10pt,conference]{ieeeconf}

\IEEEoverridecommandlockouts
\usepackage{microtype}
\usepackage{graphicx}
\usepackage{booktabs}
\usepackage{amsmath,amssymb,bm}
\usepackage{cite}
\usepackage{tabularx}
\usepackage{placeins}
\usepackage{pgfplots}
\pgfplotsset{compat=1.16}
\usepackage[hidelinks]{hyperref}

\let\citep\cite
\let\citet\cite

\newcommand{\savla}{SAVLA}
\newcommand{\groot}{GR00T N1.5}
\newcommand{\Rz}{R_z}

\title{\LARGE \bf
SAVLA: Symmetry-Aware Vision-Language-Action Models\\for Robotic Manipulation
}

\author{Junle Li$^{1}$, Weixian Waylon Li$^{1}$, Fuxiang Wu$^{2}$, Fusheng Hao$^{2}$, and Fengxiang He$^{1}$%
\thanks{$^{1}$School of Informatics, University of Edinburgh, Scotland. 
        {\tt\small junle8344@gmail.com, waylon.li@ed.ac.uk, fhe@ed.ac.uk}}%
\thanks{$^{2}$Shenzhen Institutes of Advanced Technology, Chinese Academy of Sciences, China.
        }%
}

\begin{document}

\maketitle
\thispagestyle{empty}
\pagestyle{empty}

\begin{abstract}
Vision-language-action (VLA) models have become the dominant paradigm for language-conditioned robot manipulation.
However, although images and language instructions inherently encode geometric information, VLAs acquire their spatial competence purely from demonstrations.
As a result, they are reliable only within the range of scene poses that the demonstrations cover.
We propose \savla{}, an end-to-end symmetry-aware VLA model for robust and data-efficient policy learning.
Our approach keeps the pretrained vision-language backbone entirely frozen while combining it with an equivariant flow-matching action head and a learned canonicalizer.
The head decomposes its state, action, and conditioning inputs into invariant and equivariant channels, and preserves this typing throughout all of its layers.
The canonicalizer transforms oblique-view images into a canonical frame and rotates the geometric conditions consistently.
We evaluate our model on LIBERO.
Compared with the \groot{} baseline, \savla{} improves the success rate averaged over all four LIBERO suites by $5.1$ points and increases the mean success rate under rotation on LIBERO-Goal from $41.5\%$ to $90.4\%$.
\end{abstract}

\section{Introduction}
\label{ch:intro}

Vision-language-action (VLA) models, which couple a large pretrained vision-language model with a learned action head, have quickly become the dominant recipe for language-conditioned robot manipulation.
VLAs provide a unified framework for mapping visual observations and natural-language instructions to robot actions \citep{zitkovich2023rt2,kim2024openvla,black2025pi0,bjorck2025gr00tn1}.
However, standard VLAs encode no explicit architectural prior for rotational symmetry in visuomotor control \citep{kim2025openvlaoft,shukor2025smolvla,nvidia2025gr00tn15}, and consequently remain sensitive to controlled changes in object layout, camera viewpoint, and robot initial state \citep{fei2025liberoplus}.
Recent work has enhanced VLA spatial reasoning by injecting egocentric 3D position encodings and adaptive spatial action grids \citep{qu2025spatialvla}, but the resulting policy is not explicitly constrained to be rotational equivariant.
Other works build geometric symmetry directly into generative robot policies using point-cloud observations \citep{yang2024equibot,tie2025etseed}, but their reliance on 3D point clouds limits their direct integration with VLAs built primarily on pretrained RGB encoders.

In this work, we propose \savla{}, a symmetry-aware VLA that combines a frozen, non-equivariant vision-language backbone with an equivariant action head.
\savla{} treats the output features from the frozen backbone as invariant and combines them with equivariant geometric conditions in an equivariant flow-matching action head.
It further canonicalizes oblique-view RGB images while applying the corresponding rotation to the geometric conditions.

Our main contributions are as follows:
\begin{enumerate}
\item \textbf{An equivariant action head.} We design a flow-matching action head that is equivariant.
\item \textbf{A typed interface.} We introduce an interface that generates the invariant and equivariant conditions for the action head, allowing it to be driven by a backbone that is not equivariant.
\item \textbf{An end-to-end symmetry-aware VLA.} Together with a canonicalization head that maps the oblique view into a canonical frame, these components form an end-to-end symmetry-aware VLA.
\end{enumerate}
We evaluate \savla{} on LIBERO.
Compared with the \groot{} baseline, \savla{} achieves a $5.1$-point improvement in average success rate averaged across four LIBERO suites and a substantial gain under rotation, with the mean success rate on LIBERO-Goal increasing from $41.5\%$ to $90.4\%$.
Code is available at \url{https://github.com/EVIEHub/SAVLA}.

\section{Related Work}
\label{ch:background}

\subsection{Vision-Language-Action Models}

Early VLAs, including RT-1, RT-2, and OpenVLA, generate discretized action tokens autoregressively \citep{brohan2023rt1,zitkovich2023rt2,kim2024openvla}.
More recent systems adopt continuous action modules: OpenVLA-OFT introduces parallel action decoding and action chunking \citep{kim2025openvlaoft}.
Octo provides a closely related frozen-trunk, replaceable-head factorization \citep{octo2024}.
$\pi_0$, SmolVLA, and GR00T employ diffusion or flow-matching action experts \citep{black2025pi0,shukor2025smolvla,bjorck2025gr00tn1,nvidia2025gr00tn15}.
Specifically, GR00T N1 \citep{bjorck2025gr00tn1} formalizes the split as a dual-system architecture: a vision-language model provides slow semantic reasoning, while a diffusion-transformer action head denoises action chunks at control rate, conditioned on the backbone's tokens.
\groot{} \citep{nvidia2025gr00tn15} upgrades that backbone to a SigLIP2 vision encoder paired with a Qwen3-1.7B language model, using the hidden features from the 12th layer.
\savla{} is built on \groot{}.
EquiVLA \citep{ha2026equivla} combines token-level frame averaging with a steerable flow-matching action head.
In contrast, \savla{} canonicalizes oblique-view images and conditions its $\mathrm{SO}(3)$-equivariant action head on approximately invariant VLM features and equivariant geometric reference points.

\subsection{Diffusion and Flow-Matching Policies}

Behavior cloning with a unimodal regression loss cannot represent the multimodality of demonstration data.
Diffusion Policy \citep{chi2023diffusion} addresses this by generating action chunks with a denoising diffusion model \citep{ho2020denoising}, and becomes the standard action head for imitation learning.
Flow matching \citep{lipman2023flow} offers a closely related but simpler formulation, learning a velocity field that transports noise to data.
With the optimal-transport path, the resulting trajectories are close to straight, allowing accurate sampling in few integration steps.

A parallel line of work equips such generative policies with symmetry.
Equivariant Diffusion Policy \citep{wang2024equidiff} constrains the denoising network of a diffusion policy to be $\mathrm{SO}(2)$-equivariant, reporting substantial gains in the low-data regime.
EquiBot \citep{yang2024equibot} extends the construction to $\mathrm{SIM}(3)$ equivariance on point-cloud observations, covering scale in addition to rotation and translation.
Several works target the full $\mathrm{SE}(3)$ group.
Diffusion-EDFs \citep{ryu2024diffusionedf} formulates pick-and-place as bi-equivariant denoising generative modeling on $\mathrm{SE}(3)$.
RiEMann \citep{gao2024riemann} learns $\mathrm{SE}(3)$-equivariant manipulation directly from unsegmented point clouds at near real-time rates.
ET-SEED \citep{tie2025etseed} lifts the equivariance from single poses to whole trajectories.

\subsection{Canonicalization}

An alternative to building equivariance into the architecture is to normalize the input instead.
Learned canonicalization \citep{kaba2023equivariance} trains a small network to predict a canonical pose.
The input is mapped into that frame before the backbone, and the output is mapped back afterwards.
Probabilistic symmetrization \citep{kim2023learning} generalizes the idea to sampled group elements.
In robot manipulation, Canonical Policy \citep{zhang2025canonical} learns a canonical 3D representation for an $\mathrm{SE}(3)$-equivariant visuomotor policy.

The effectiveness of equivariance depends on the accuracy of the canonicalizer: errors in the estimated pose propagate directly to the prediction.
Architectural equivariance offers the opposite trade-off: it guarantees equivariance by construction, but requires equivariant design throughout the network, which is impractical for a multi-billion-parameter pretrained backbone.
\savla{} deliberately combines the two approaches at different stages of the pipeline.
Canonicalization normalizes the input image for the frozen backbone, whose features are subsequently treated as invariant, while the estimated scene-rotation is also used to rotate the geometric conditions of the equivariant action head (Section~\ref{sec:interface}).

\section{Preliminaries}
\label{ch:prelim}
\label{sec:prelim}

\paragraph*{Problem setup}
We consider language-conditioned manipulation.
At each control step $t$, the policy receives an observation $o_t=(I_t^{3p},I_t^w,s_t,\ell)$, where $I_t^{3p}$ is the third-person image, $I_t^w$ is the wrist image, $s_t$ is the proprioceptive state of the arm, and $\ell$ is the language instruction.
The policy then generates a chunk of future actions \((a_t,\ldots,a_{t+T_p-1})\), where $T_p$ is the horizon.

\paragraph*{Equivariance and typed features}
A group $G$ acts on a vector space through a representation $\rho$, which assigns to each $R \in G$ an invertible linear map $\rho(R)$.
A map $f$ is equivariant if transforming its input by a group element produces the correspondingly transformed output: $f(\rho_{\text{in}}(R)x) = \rho_{\text{out}}(R)f(x)$; invariance is the special case: $f(\rho_{\text{in}}(R)x) = f(x)$.
Any finite-dimensional representation of a compact group can be decomposed into a direct sum of irreducible representations (irreps), which define the basic transformation types of the features.
The trivial irrep $\rho_0$, for which $\rho_0(R) = 1$, is a scalar unchanged by rotation (a type-0 track); the standard irrep $\rho_1$, for which $\rho_1(R) = R$, is a vector in $\mathbb{R}^3$ that co-rotates with the scene (a type-1 track).
Higher-order irreps exist but are not used here.
It is worth noting that, while the action head is exactly $\mathrm{SO}(3)$-equivariant, the canonicalizer operates on gravity-axis rotations $R_z(\theta)$, with ${R_z(\theta)}\cong\mathrm{SO}(2)\subset\mathrm{SO}(3)$.
Accordingly, our end-to-end equivariance concerns this subgroup.

\paragraph*{Flow matching}
The action chunk is generated by a conditional model trained with flow matching \citep{lipman2023flow}, which learns a velocity field that transports noise toward the data distribution along near-straight paths.
Given a ground-truth chunk $A_t$, a noise sample $\epsilon \sim \mathcal{N}(0, I)$ and a flow time $\tau \in [0,1]$, we form the interpolant $Z_\tau=\tau A_t+(1-\tau)\epsilon$ and train $v_\phi(Z_\tau,S_t,\tau,\mathcal{C}_t)$ to predict the constant velocity $A_t-\epsilon$ that connects the two endpoints:
\begin{equation}
  \mathcal{L} \;=\; \mathbb{E}_{A_t, \epsilon, \tau}
  \big\| v_\phi(Z_\tau, S_t, \tau, \mathcal{C}_t)
  - (A_t - \epsilon) \big\|^2 ,
  \label{eq:fm}
\end{equation}
where $S_t$ is the typed state representation derived from $s_t$ and $\mathcal{C}_t$ is the condition of the flow velocity field.
At inference we sample $Z_0=\epsilon\sim\mathcal{N}(0,I)$ and integrate the learned velocity field with $K$ steps,
\begin{equation}
  Z_{k+1}=Z_k+\tfrac{1}{K}\,
  v_\phi(Z_k,S_t,\tau_k,\mathcal{C}_t),
  \qquad \tau_k = k/K,
\end{equation}
then $Z_K$ is the predicted action chunk.

Let $\rho_A$, $\rho_S$, and $\rho_C$ be the rotation representations on action chunks, states, and conditions.
For any $R\in\mathrm{SO}(3)$, the design target is
\begin{equation}
  \begin{aligned}
    v_\phi\!\big(
      \rho_A(R)Z_\tau,\rho_S(R)S_t,\tau,
      &\rho_C(R)\mathcal{C}_t
    \big) \\
    &=\rho_A(R)v_\phi(Z_\tau,S_t,\tau,\mathcal{C}_t).
  \end{aligned}
  \label{eq:equivariance}
\end{equation}

\section{Method}
\label{ch:method}

\begin{figure*}[t]
\centering
\includegraphics[width=\textwidth]{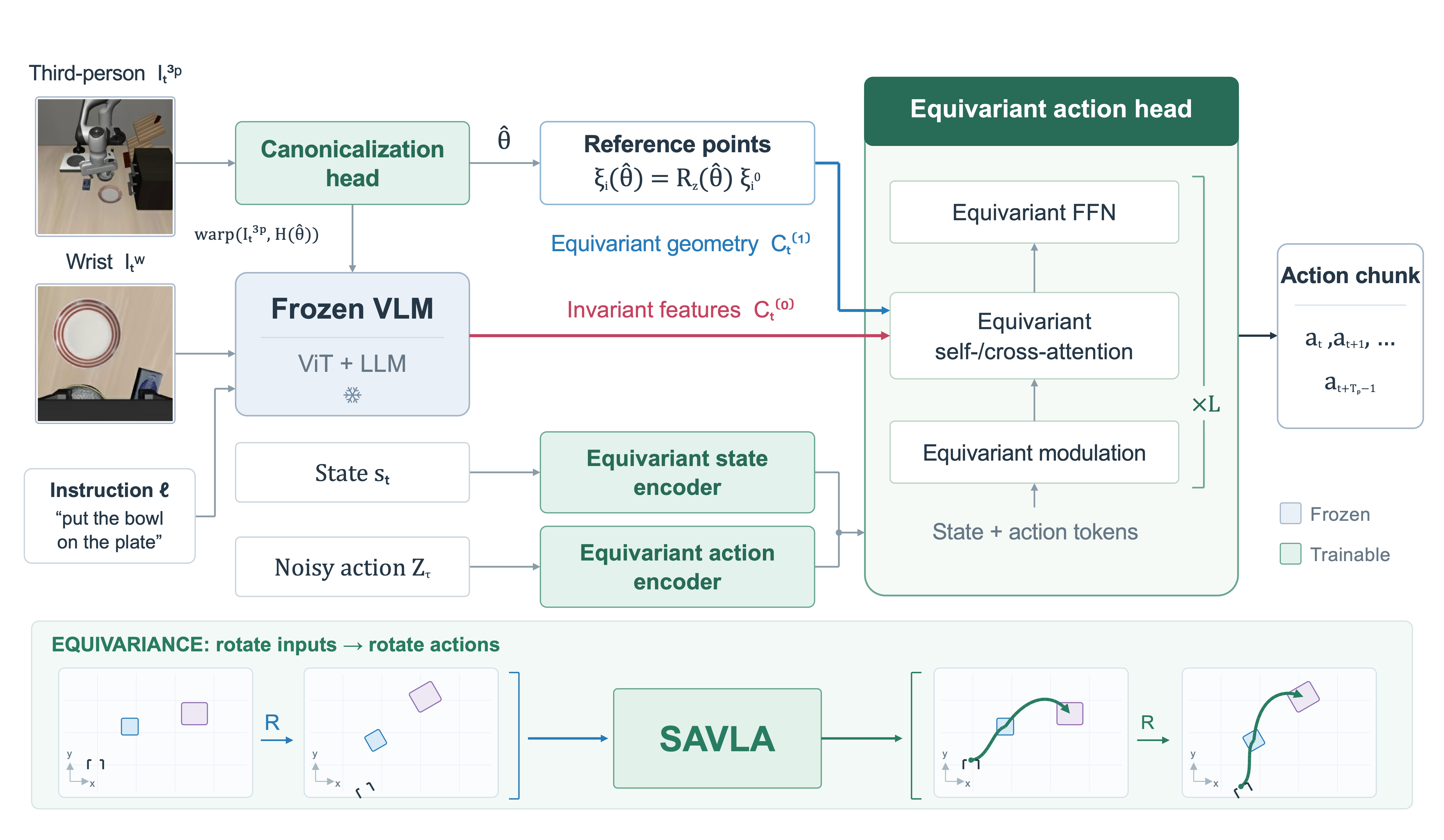}
\caption{\savla{} architecture.
The canonicalization head estimates the scene rotation $\hat\theta$ from the third-person image, constructs the canonical view $\operatorname{warp}(I_t^{3p},H(\hat\theta))$ for the frozen VLM, and rotates the fixed reference points.
Equivariant encoders embed the state and the noisy action.
Equivariant flow-matching head outputs the action chunk.}
\label{fig:arch}
\end{figure*}

In this section, we will describe the design of \savla{}.
Our method builds upon VLA policies that combine a pretrained vision-language backbone with a flow-matching action head \citep{nvidia2025gr00tn15}.
Starting from this architecture, our key insight is that equivariance can be introduced without modifying the frozen, non-equivariant backbone.
We treat the backbone tokens as type-0 conditions, construct type-1 geometric conditions from a set of fixed reference points.
By building the flow-matching velocity network entirely from symmetry-preserving operations, we make every integration update equivariant by construction.
To handle RGB observations captured from an oblique viewpoint, \savla{} uses a learned canonicalizer that maps the third-person image into a canonical view and rotates the geometric conditions consistently. Figure~\ref{fig:arch} gives an overview of the full architecture.


\subsection{Equivariant Action Head}
\label{sec:head}
The flow-matching sampler starts from an isotropic Gaussian action chunk $Z_0\sim\mathcal{N}(0,I)$ and transforms it into $Z_K$ through $K$ explicit Euler updates.
The distribution of $Z_0$ is invariant under $\rho_A(R)$.
Given this invariant initialization, if the learned velocity field satisfies Eq.~\eqref{eq:equivariance}, the resulting flow preserves equivariance throughout the integration.
Let $Z'_k=\rho_A(R)Z_k$ be the corresponding rotated trajectory, then
\begin{equation}
  \begin{aligned}
  Z'_{k+1}
  &= \rho_A(R)Z_k + \tfrac{1}{K}
  v_\phi\!\left(\rho_A(R)Z_k,\rho_S(R)S_t,\tau_k,
  \rho_C(R)\mathcal{C}_t\right) \\
  &= \rho_A(R)\!\left[Z_k + \tfrac{1}{K}
  v_\phi(Z_k,S_t,\tau_k,\mathcal{C}_t)\right]
   = \rho_A(R)Z_{k+1}.
  \end{aligned}
  \label{eq:eqflow}
\end{equation}
The rest of this section describes how each layer makes the velocity field satisfy Eq.~\eqref{eq:equivariance}.

To construct such an equivariant velocity field, each hidden token in the velocity network is split into a scalar track $h\in\mathbb{R}^{C_0}$ and a vector track $V=(v_1,\ldots,v_{C_1})\in\mathbb{R}^{C_1\times 3}$, where $C_0$ and $C_1$ denote the numbers of type-0 and type-1 channels.
Following the definitions in Section~\ref{sec:prelim}, $h$ is invariant under rotation, whereas every $v_c\in\mathbb{R}^3$ co-rotates with the input.

\paragraph*{Action and state representation}
Before describing the internal layers, we first specify how the state and action are mapped to these two transformation tracks.
The raw $8$-D state contains end-effector position $p$, absolute orientation as an axis--angle vector, and two gripper joint positions $s_{\mathrm{grip}}$.
We center position about the workspace rotation center $c_{\mathrm{ws}}$, convert orientation to a rotation matrix $R_{\mathrm{ee}}=[r_1\;r_2\;r_3]\in\mathrm{SO}(3)$, and use its first two columns $r_1,r_2$.
We also append the known gravity direction $g_{\mathrm{grav}}=(0,0,1)$.
Thus the typed state is $[p-c_{\mathrm{ws}},r_1,r_2,g_{\mathrm{grav}}]\oplus s_{\mathrm{grip}}$, with signature $4\rho_1\oplus2\rho_0$.
The action $a_t=(\Delta p_t,\omega_t,u_t^{\mathrm{grip}})$ comprises a world-frame position increment, an incremental axis--angle rotation vector and a scalar gripper command.
It has signature $2\rho_1\oplus\rho_0$.

The $\rho_0$ components are embedded by ordinary MLPs, while $\rho_1$ components are embedded by Vector Neuron \citep{deng2021vector} $V'=WV$, which operate only the channel dimension, leaving the three spatial components unchanged.

The resulting state and action embeddings form the input tokens of the velocity network.
They are processed by $L$ repeated blocks, each consisting of a normalization and modulation module, an attention module, and an FFN module.
Next, we will show how these operations preserve the scalar and vector transformation types.

\paragraph*{Normalization and modulation}
Each block first normalizes and modulates its input while maintaining the transformation behavior of each track.
The scalar track uses LayerNorm, whereas the vector track uses an equivariant RMS norm: let $q=\mathrm{RMS}_{m'}\big(\|v_{m'}\|\big)$ be its invariant scale, then the normalized vector feature is $\tilde v_m=v_m/q$.
For modulation, we use an equivariant modulation that changes the magnitude of each feature vector without altering its direction.
Let $g=(g_1,\ldots,g_{C_1})\in\mathbb{R}^{C_1}$ be the vector of gates produced by an ordinary MLP of the flow-timestep embedding.
For each channel $m$, the modulated vector output is
\begin{equation}
  v'_m \;=\; (1 + g_m)\, \tilde v_m.
  \label{eq:eqmod}
\end{equation}

The scalar track uses standard modulation.
Let $\gamma_0,\beta_0\in\mathbb{R}^{C_0}$ be the modulation parameters produced by an MLP of the flow-timestep embedding.
The modulated scalar output is:
\begin{equation}
  h' \;=\; \mathrm{LN}(h) \odot (1 + \gamma_0) \;+\; \beta_0 ,
  \label{eq:scalarmod}
\end{equation}

\paragraph*{Equivariant attention}
The action head uses self-attention among the input tokens (state tokens + action tokens) and cross-attention from the input tokens to the condition tokens.
The state and action tokens are typed by their respective encoders, and the condition tokens are typed by the interface of Section~\ref{sec:interface}.
The Q/K/V projections are applied per track: an ordinary linear layer on the scalar part, a Vector Neuron layer on the vector part.
Each query and key carries both a scalar and a vector component.
A rotation acts on these features through a block-diagonal orthogonal transformation, which preserves the inner products $\langle q,k\rangle$ and hence leaves the attention weights unchanged.
We sum them into a single attention weight:
\begin{equation}
  \begin{gathered}
  \mathrm{score}_{ij} \;=\;
  \frac{\langle q^{s}_i, k^{s}_j \rangle}{\sqrt{d_s}}
  \;+\; \lambda_h \sum_m
  \frac{\langle q^{V,m}_i, k^{V,m}_j \rangle}{\sqrt{3d_v}}, \\
  \alpha_{ij} \;=\; \mathrm{softmax}_j\big(\mathrm{score}_{ij}\big),
  \end{gathered}
  \label{eq:eqattn}
\end{equation}
where $d_s=C_0/N_h$, $d_v=C_1/N_h$ are the scalar-channel and vector-channel widths per head, $N_h$ is the number of attention heads, $m=1,\ldots,d_v$, and $\lambda_h$ is a learnable parameter for head $h$.
Since $\alpha_{ij}$ is invariant, aggregation with the same $\alpha$ on both tracks returns an invariant scalar output and an equivariant vector output: an invariant-weighted sum of equivariant values is equivariant.
In self-attention both sides are input tokens; in cross-attention the keys and values come from the condition.
Multi-head splitting operates along the channel dimension.
Since scalar and vector channels are partitioned across heads, each head remains equivariant.

\paragraph*{Equivariant FFN}
After attention, each token is processed by a pointwise FFN.
The scalar track uses a standard MLP with GELU, but applying the same nonlinearity to the vector track would break equivariance: $\mathrm{GELU}(R v) \neq R\,\mathrm{GELU}(v)$.
To apply a GELU on the vector track without breaking equivariance, we use an equivariant neural network to produce a per-channel direction.
Let $W_k$ be the equivariant neural network, then $k_c=(W_kV)_c$ is the direction for channel $c$, and $\hat k_c=k_c/\|k_c\|$ is the corresponding unit vector.
Each feature vector is decomposed into components parallel and perpendicular to $\hat k_c$, and only the parallel part passes through the GELU while the perpendicular component remains unchanged:
\begin{equation}
  v'_c \;=\; \mathrm{GELU}\big(\langle v_c, \hat k_c \rangle\big)\,
  \hat k_c
  \;+\; \big( v_c - \langle v_c, \hat k_c \rangle\, \hat k_c \big).
  \label{eq:eqffn}
\end{equation}
Since $k_c \mapsto Rk_c$ under rotation and $R$ preserves both $\|k_c\|$ and the inner product $\langle v_c, \hat{k}_c \rangle$, the parallel and perpendicular components in Eq.~\eqref{eq:eqffn} transform equivariantly under $R$.
The complete FFN applies this activation between the up- and down-projections.

\paragraph*{Cross-track interaction}
Although the two tracks follow different transformation rules, they can still exchange information through symmetry-preserving cross-track operations: vector $\to$ scalar through the per-channel invariant norms $\mathrm{Inv}(V)$, concatenated into the scalar track at every block; scalar $\to$ vector through invariant coefficients, which rescale equivariant quantities without altering their directions.

Since each layer preserves the transformation law, the entire head satisfies Eq.~\eqref{eq:equivariance}.

An equivariant action head alone, however, is insufficient to make the complete VLA equivariant, because the frozen vision-language backbone has no prescribed transformation law under rotations.
Sections~\ref{sec:interface} and~\ref{sec:canon} therefore introduce a typed conditioning interface and a learned canonicalizer to extend the head-level equivariance guarantee to approximate end-to-end rotational equivariance.

\subsection{Symmetry-aware Conditions}
\label{sec:interface}

The equivariance guarantee of Section~\ref{sec:head} assumes that the condition obeys the transformation law.
Since the frozen VLM does not impose such a transformation law on its output tokens, we define a typed interface that provides the action head with invariant and equivariant conditioning tracks.
The construction below assumes that the canonicalizer provides a canonical third-person image and a scene-rotation estimate $\hat\theta$; Section~\ref{sec:canon} describes how these quantities are obtained.

\paragraph*{Invariant tracks}
Since the the third-person image is canonicalized, with the wrist image and language instruction remaining invariant to scene rotations, the resulting VLM features are approximately invariant to such rotations.
Let $N_c$ denote the total number of condition tokens and $D_0$ the scalar-channel width.
Accordingly, the scalar conditioning track is represented as $C_t^{(0)}\in\mathbb{R}^{N_c\times D_0}$.

\paragraph*{Equivariant tracks}
A set of fixed reference points $X=\{\boldsymbol{\xi}_i^0\}_{i=1}^{N_{3p}}$ is precomputed by back-projecting the ViT patch centers through the camera onto the tabletop plane, where $N_{3p}$ is the number of third-person visual tokens.
Each $\boldsymbol{\xi}_i^0$ is rotated by the scene rotation estimate $\hat\theta$ from the canonicalizer:
\begin{equation}
  \boldsymbol{\xi}_i(\hat\theta)
  \;=\; \Rz(\hat\theta)\,\boldsymbol{\xi}_i^0,
  \label{eq:anchors}
\end{equation}
where $R_z(\hat\theta)\in\mathrm{SO}(2)$ is the rotation by $\hat\theta$ about the gravity axis.
Each third-person visual token is therefore paired with an equivariant location.
Aligning the vector conditioning track with the scalar one, we define
\begin{equation}
  \big(C_t^{(1)}\big)_{j}
  =
  \begin{cases}
    \boldsymbol{\xi}_i(\hat\theta),
      & \text{\shortstack[l]{if token $j$ is the $i$th\\
                              third-person visual token}}\\[0.6em]
    \boldsymbol{0},
      & \text{\shortstack[l]{if token $j$ is a wrist-camera\\
                              or language token}}
  \end{cases}
  \label{eq:vector-condition}
\end{equation}

Together, the two tracks form the complete condition of the flow velocity field, $\mathcal{C}_t=(C_t^{(0)},C_t^{(1)})$.
At interface, we define its group action as $\rho_C(R)\mathcal{C}_t=(C_t^{(0)},R C_t^{(1)})$.
\subsection{Canonicalization}
\label{sec:canon}

Section~\ref{sec:interface} requires the third-person observation to be mapped to an approximately rotation-invariant canonical view, while geometric reference points co-rotate with the scene.
Canonicalization satisfies these requirements by using the same estimated rotation to warp the image and transform the reference points.
In most manipulation setups, the third-person camera observes the table from an oblique rather than a top-down viewpoint.
Under an oblique view, rotating the scene about the gravity axis is not an in-plane image rotation, and naively rotating the image about its center misaligns the scene.
To present the frozen backbone with a canonical view, we therefore warp the image by the homography induced by the dominant tabletop plane \citep{hartley2004multiple}.
Let $H(\theta)$ denote the homography from the canonical image to the image obtained after a scene rotation $\theta$:
\begin{equation}
  H(\theta) \;=\; K_{\mathrm{cam}} \Big( R_\theta
  + \tfrac{1}{d}\, t_\theta\, n^\top \Big) K_{\mathrm{cam}}^{-1},
  \label{eq:homography}
\end{equation}
where $K_{\mathrm{cam}}$ is the camera intrinsic matrix, $(R_\theta,t_\theta)$ is the relative camera motion equivalent to the scene rotation, and $(n,d)$ parameterizes the tabletop plane in the camera frame.
These quantities are computed from the simulator camera specification and table geometry rather than estimated from demonstrations.
We denote by $\operatorname{warp}(x,H)$ the inverse-sampling image warp defined by $\operatorname{warp}(x,H)(u)=x(Hu)$, i.e., each output pixel at $u$ samples the source image at the corresponding location $Hu$.
Under this convention, applying $\operatorname{warp}(x_\theta,H(\theta))$ to an observation at rotation $\theta$ maps its content back to the canonical view.

To get the scene rotation estimate $\hat\theta$, we use a three-layer CNN with global average pooling and a linear head ($0.09$ million parameters in total) to score $M=8$ candidates $\theta_k=2\pi k/M$, for $k=0,\ldots,M-1$.
For each candidate, the canonical view is constructed as
\begin{equation}
  x_k=\operatorname{warp}\!\left(x,H(\theta_k)\right).
  \label{eq:canon-candidate}
\end{equation} 
Let $p_k$ be the softmax probability of candidate $\theta_k$.
The continuous estimate $\hat\theta$ is obtained by the circular mean
\begin{equation}
  \hat\theta=\operatorname{atan2}\!\left(\sum_k p_k\sin \theta_k,
  \sum_k p_k\cos \theta_k\right).
  \label{eq:canon-mean}
\end{equation}
The scorer is trained in a self-supervised manner using the original demonstrations.
Each demonstration frame $x$ is treated as the canonical pose.
We sample a rotation offset $\Delta\sim\mathcal{U}(-\pi,\pi)$ and generate a synthetic frame $x_\Delta=\operatorname{warp}\!\left(x,H(\Delta)^{-1}\right)$.
Let $\hat\theta_\Delta$ denote the angle predicted from the
synthetically rotated frame $x_\Delta$, while $\hat\theta_0$
is the prediction from the original demonstration frame $x$.
We train the scorer with
\begin{equation}
\begin{aligned}
\mathcal{L}_{\mathrm{canon}}
= \mathbb{E}_{x,\Delta}\Big[
& (1-\cos\hat\theta_0) \\
& + \lambda\big(1-\cos(
\hat\theta_\Delta-\hat\theta_0-\Delta)\big)
\Big],
\end{aligned}
\end{equation}
In the second term, gradients through $\hat\theta_0$ are stopped.
To calibrate the scale of the estimated rotation magnitude, we then introduce a scalar gain using $N_{\mathrm{cal}}=12$ simulator frames rendered at known angles, which are disjoint from the evaluation set.
At inference, the canonicalized third-person image $\operatorname{warp}(x,H(\hat\theta))$ is passed to the frozen VLM, while the same estimate $\hat\theta$ rotates the fixed reference points in Eq.~\eqref{eq:anchors}.

Consequently, canonicalization makes the third-person image approximately invariant under rotations, while the rotated reference points provide equivariant geometric conditions.
Together with the equivariant action head, \savla{} achieves approximate end-to-end rotational equivariance.

\section{Experiments}
\label{ch:experiments}

Our experiments are designed to evaluate \savla{} along three main dimensions: performance on standard language-conditioned manipulation tasks, robustness to unseen scene rotations, and data efficiency under reduced demonstration budgets.
We use all four LIBERO suites as the primary benchmark.
First, we assess standard task performance, followed by an evaluation of rotation generalization under rotations.
To better understand the rotation result, we compare against rotation augmentation, ablate the equivariant action head and canonicalization module, and use oracle variants to locate the remaining error.
In addition, we study data efficiency by reducing the available demonstrations.
Finally, to assess its effectiveness in real-world settings, we evaluate \savla{} on three real-world manipulation tasks. Since \savla{} is built on \groot{}, we use a budget-matched \groot{} as our primary baseline.

For a controlled comparison, \savla{} and the \groot{} baseline use the same frozen vision-language backbone, demonstrations, and checkpoint-selection procedure, with both action heads trained from scratch. 
\savla{} uses $L=16$ blocks, typed widths $C_0=C_1=256$, $8$ attention heads, an action horizon $T_p=16$, and $K=4$ flow-integration steps.
To match the parameter count of \savla{}, we reduce the official \groot{} flow-matching head to 7 attention heads with a hidden width of 192 and use a single-layer VLM-feature self-attention encoder, yielding 100.8M trainable parameters compared with 103.3M for \savla{}.
Both models are optimized with AdamW using a learning rate of $10^{-4}$ and a batch size of $16$.
For each suite, checkpoints at $\{20\mathrm{K},40\mathrm{K},80\mathrm{K}\}$ steps are screened using the same $200$-episode preliminary evaluation, and the best checkpoint is used for reporting.
The canonicalizer is trained separately for $3{,}000$ steps using AdamW with a learning rate of $10^{-4}$ and a batch size of $32$.

\subsection{LIBERO Experimental Results}
\label{sec:exp-indist}

\begin{table}[b]
\centering
\caption{Experimental results on LIBERO. External baseline results are taken from the original papers or subsequent published evaluations.}
\label{tab:main}
\resizebox{\columnwidth}{!}{%
\begin{tabular}{lccccc}
\toprule
Method & Spatial & Object & Goal & Long & Average \\
\midrule
Diffusion Policy~\citep{chi2023diffusion} & 78.3 & 92.5 & 68.3 & 50.5 & 72.4 \\
Octo~\citep{octo2024}                     & 78.9 & 85.7 & 84.6 & 51.1 & 75.1 \\
OpenVLA~\citep{kim2024openvla}            & 84.7 & 88.4 & 79.2 & 53.7 & 76.5 \\
SmolVLA~\citep{shukor2025smolvla}         & 93.0 & \textbf{94.0} & 91.0 & 77.0 & 88.8 \\
\midrule
\groot{}            & 95.3 & 92.4 & 91.6 & 66.7 & 86.5 \\
\savla{} (ours)                           & \textbf{96.7} & 92.9 & \textbf{97.5} & \textbf{79.3} & \textbf{91.6} \\
\bottomrule
\end{tabular}
}
\end{table}

\begin{table}[t]
\centering
\caption{Success rates (\%) on rotation augmentation variants.}
\label{tab:aug}
\resizebox{\columnwidth}{!}{%
\begin{tabular}{lcc}
\toprule
Method & $-30^\circ$, \dots, $+30^\circ$\ & $\pm 35/45^\circ$ \\
\midrule
\groot{}                          & 41.5 & 1.9 \\
\groot{} + rotation aug.\ (warped)      & 69.1 & 24.0 \\
\groot{} + rotation aug.\ (re-rendered) & 74.5 & 59.2 \\
\savla{} (ours)                   & \textbf{90.4} & \textbf{64.5} \\
\bottomrule
\end{tabular}}
\end{table}

We evaluate SAVLA on all four LIBERO suites. Each model is evaluated on $500$ episodes per suite, and we report the mean over three random seeds.
As shown in Table~\ref{tab:main}, \savla{} outperforms the \groot{} baseline on all four suites, increasing the average success rate from $86.5\%$ to $91.6\%$ ($+5.1$ points).
The results indicate that the symmetry-aware architecture can improve standard in-distribution manipulation performance.
The largest gains occur on Goal and Long, with improvements of $5.9$ and $12.6$ points.
The consistently lower scores on Long across methods reflect its longer, multi-stage task horizons, over which execution errors can accumulate \citep{liu2023libero}.
We exclude some baselines that use substantially larger models or additional pretraining on robot data to improve comparability.
\subsection{Rotation Generalization}
\label{sec:exp-rotgen}

\paragraph*{Rotation Robustness}
\label{sec:exp-rot}

\begin{figure*}[t]
\centering
\includegraphics[width=0.96\textwidth]{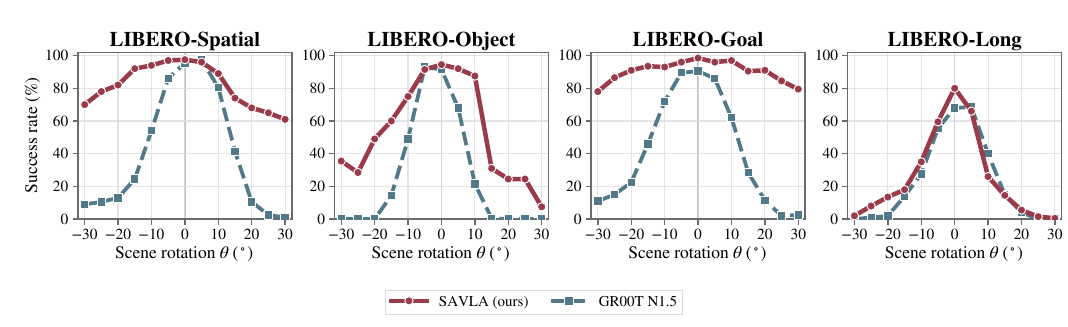}
\caption{Rotation-OOD across all four LIBERO suites.
Each panel compares \savla{} with \groot{} over the same scene-rotation sweep.
Mean over the 13 angles (\savla{} vs.\ \groot{}): Spatial 81.8\% vs.\ 40.5\%, Object 53.9\% vs.\ 26.0\%, Goal 90.4\% vs.\ 41.5\%, and Long 25.4\% vs.\ 22.7\%.}
\label{fig:rotation}
\end{figure*}

We analyze rotation robustness on LIBERO by rotating the entire scene (the table layout, the objects, and the robot end-effector state) about the gravity axis through the workspace center by $\theta \in \{-30^\circ, \dots, +30^\circ\}$ in $5^\circ$ steps during evaluation.
This range introduces unseen scene orientations while keeping the task semantics, robot reachability, and camera visibility comparable to the original benchmark.
We implement the scene rotation by orbiting the third-person camera by $-\theta$, transforming the proprioceptive state into the rotated frame, and mapping the predicted actions back to the original simulator frame before execution.
Both methods are trained only on the original dataset.
At each of the $13$ scene rotations, we evaluate $200$ episodes using three random seeds.
Figure~\ref{fig:rotation} shows that \savla{} substantially improves rotation robustness on Spatial, Object, and Goal, increasing the mean success rates over the 13 angles from $40.5\%$ to $81.8\%$, $26.0\%$ to $53.9\%$, and $41.5\%$ to $90.4\%$, respectively.
The smaller improvement on Long ($22.7\%$ to $25.4\%$) may be attributed to its longer, multi-stage horizon, which compounds execution errors beyond what rotational equivariance alone can address.
We conduct the following augmentation and ablation studies on LIBERO-Goal as a representative suite.

\paragraph*{Rotation Augmentation}
\label{sec:exp-aug}

A natural alternative to building symmetry into the architecture is to learn it from data.
Both augmentation variants consistently transform the third-person image, proprioceptive state, and actions.
In the warped variant, we sample $\theta \sim \mathcal{U}(-30^\circ, +30^\circ)$ for each training sample and warp the third-person image using the homography in Eq.~\eqref{eq:homography}.
In the re-rendered variant, we render the third-person image from rotated camera viewpoints at the same $13$ discrete angles used in the rotation experiment.
Both variants are trained with the same recipe and budget as the unaugmented baseline.

Table~\ref{tab:aug} shows that warped rotation augmentation improves the unaugmented \groot{} baseline across both reported metrics.
The re-rendered variant further improves rotation robustness, reaching $74.5\%$ over $-30^\circ$ to $30^\circ$ and $59.2\%$ at the extrapolation angles.
Nevertheless, \savla{} achieves higher success rates of $90.4\%$ and $64.5\%$ without additional rotated demonstration trajectories, highlighting the advantage of our symmetry-aware design over rotation augmentation in this setting.


\subsection{Ablation Study}
\label{sec:exp-ablation}

\begin{figure}[t]
\centering
\includegraphics[width=\columnwidth]{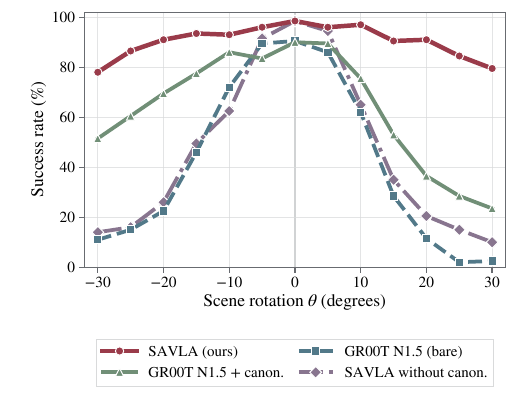}
\caption{Ablation on the LIBERO-Goal rotation experiment.
Means over the 13 angles: \savla{} 90.4, \savla{} without canonicalization 46.0, GR00T N1.5 + canonicalization (full) 63.5, GR00T N1.5 41.5.}
\label{fig:ablation}
\end{figure}

\begin{table}[t]
\centering
\caption{Real-robot success rates (\%) over $20$ episodes per task.}
\label{tab:realrobot}
\resizebox{\columnwidth}{!}{%
\begin{tabular}{lcccc}
\toprule
Method & \shortstack{Pick and\\place} & \shortstack{Block\\stacking} & \shortstack{Block\\sorting} & Avg. \\
\midrule
\groot{} & 70 & 60 & 55 & 61.7 \\
\savla{} (ours) & \textbf{75} & \textbf{70} & \textbf{70} & \textbf{71.7} \\
\bottomrule
\end{tabular}}
\end{table}

To locate where the rotation robustness comes from, we ablate both components.
For \groot{}, we apply a full canonicalization wrapper: the same canonicalizer normalizes the third-person image and provides the rotation estimate used to transform the robot state into the canonical frame.
The standard action head predicts actions in this frame, which are then transformed back to the execution frame.
Conversely, we remove the canonicalization from \savla{} by forcing $\hat\theta = 0$.
All variants in this section are evaluated over $200$ episodes with three random seeds.

Figure~\ref{fig:ablation} shows that neither component is sufficient on its own: adding canonicalization to \groot{} reaches $63.5\%$, while removing canonicalization from \savla{} yields $46.0\%$, compared with $90.4\%$ for the complete model.
The equivariant action head and canonicalization module are therefore jointly necessary for the rotation robustness.
We next use two oracle variants to determine which part of the system accounts for its remaining error.

As shown in Table~\ref{tab:oracle}, replacing the estimated rotation angle by the ground-truth angle, while leaving the warp untouched, raises the mean from $90.4$ to $90.9$.
With the true canonical view in place of the warped image, the mean rises to $97.3$ and remains at $96.8$ even when $|\theta| \geq 25^\circ$.
These results suggest that the remaining headroom on LIBERO-Goal lies primarily in front-end image geometry rather than scene rotation estimation.

\begin{table}[t]
\centering
\caption{Success rates (\%) on oracle variants.}
\label{tab:oracle}
\resizebox{\columnwidth}{!}{%
\begin{tabular}{lcc}
\toprule
Variant & mean over $13$ angles & $|\theta| \geq 25^\circ$ \\
\midrule
\savla{} (estimated rotation angle, warped image) & 90.4 & 82.1 \\
\quad + ground-truth angle               & 90.9 & 83.4 \\
\quad + true canonical view            & \textbf{97.3} & \textbf{96.8} \\
\bottomrule
\end{tabular}}
\end{table}


\subsection{Data Efficiency}
\label{sec:exp-lowdata}

\begin{table}[t]
\centering
\caption{Data efficiency on LIBERO (average success rate \% over the four suites)}
\label{tab:lowdata}
\small
\begin{tabular}{lccc}
\toprule
Method & 10\% data & 25\% data & 100\% data \\
\midrule
\groot{} & 48.3 & 66.1 & 86.5 \\
\savla{} (ours)               & \textbf{57.1} & \textbf{71.9} & \textbf{91.6} \\
\midrule
$\Delta$                      & +8.8 & +5.8 & +5.1 \\
\bottomrule
\end{tabular}
\end{table}

In addition to rotation robustness, we further investigate whether \savla{} maintains its advantage when fewer demonstrations are available.
We re-train the models and the canonicalizer on subsets of the demonstrations ($10\%$ and $25\%$ per task, fixed seed, identical subsets for both methods) and evaluate over $200$ episodes with three random seeds, while the $100\%$ result is taken from Table~\ref{tab:main}.
As shown in Table~\ref{tab:lowdata}, the gap increases from $5.1$ points with the full dataset to $5.8$ points at $25\%$ data and $8.8$ points at $10\%$ data, showing that the equivariance structure is still beneficial under a low demonstration budget.

\subsection{Real Robot Experiment}
\label{sec:exp-realrobot}

\begin{figure}[t]
\centering
\includegraphics[width=\columnwidth]{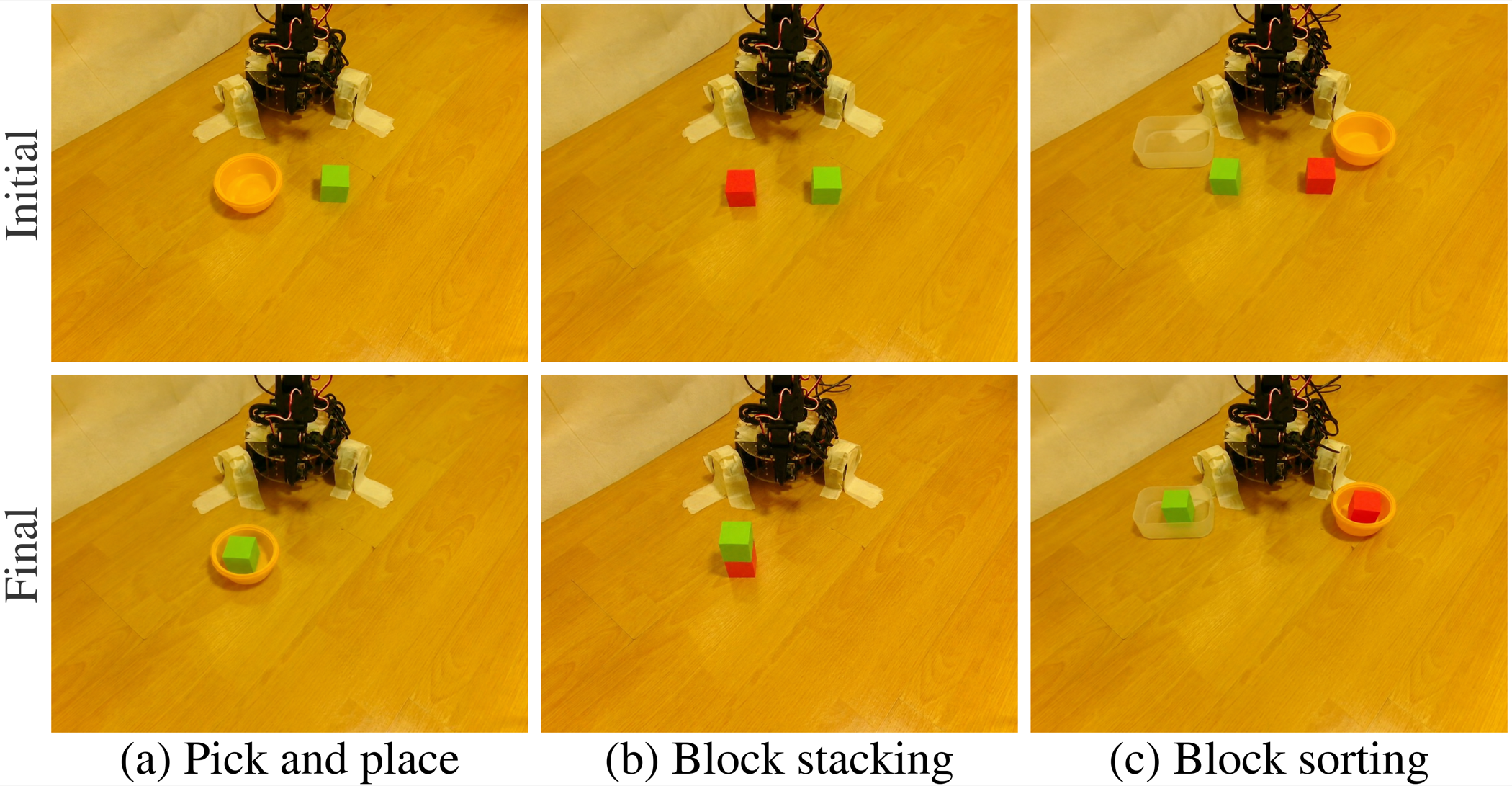}
\caption{Initial and final frame of the three real-robot tasks.}
\label{fig:realrobot}
\end{figure}

To test the real-world applicability, we evaluate \savla{} and \groot{} on the SO-101 robot across three real-robot manipulation tasks: pick-and-place, block stacking, and block sorting. 
Following the simulation setup, both models use observations from a third-person camera and a wrist camera.
Figure~\ref{fig:realrobot} shows the initial and final frame of three real-robot tasks.
We collect $50$ demonstrations per task, yielding $150$ demonstrations in total. Both models are trained in a multi-task setting on demonstrations from all three tasks for $40\mathrm{K}$ steps, using a batch size of $16$ and a learning rate of $1\times10^{-4}$.
We evaluate each task over 20 episodes and report the success rate.

As shown in Table~\ref{tab:realrobot}, \savla{} outperforms \groot{} on all three real-world tasks, increasing the mean success rate from $61.7\%$ to $71.7\%$. The result is consistent with our simulation findings, indicating that \savla{} also improves manipulation performance in real-world settings.

We further evaluate rotation robustness on pick-and-place at $\pm10^\circ$, $\pm20^\circ$, and $\pm30^\circ$, using $20$ episodes per angle.
We use the same camera-orbit and state/action coordinate transformations as in Section~\ref{sec:exp-rotgen}.
As shown in Table~\ref{tab:realrobot-rotation}, \savla{} increases the mean success rate across the seven angles from $47.9\%$ to $57.1\%$, demonstrating stronger rotation robustness in the real-world experiments.

\begin{table}[t]
\centering
\caption{Real-robot pick-and-place success rates (\%) under rotation.}
\label{tab:realrobot-rotation}
\resizebox{\columnwidth}{!}{%
\begin{tabular}{lcccccccc}
\toprule
Method & $-30^\circ$ & $-20^\circ$ & $-10^\circ$ & $0^\circ$ & $10^\circ$ & $20^\circ$ & $30^\circ$ & Avg. \\
\midrule
\groot{} & 35 & 45 & 55 & 70 & 55 & 45 & 30 & 47.9 \\
\savla{} (ours) & \textbf{45} & \textbf{55} & \textbf{60} & \textbf{75} & \textbf{65} & \textbf{55} & \textbf{45} & \textbf{57.1} \\
\bottomrule
\end{tabular}}
\end{table}

\section{Conclusion}
\label{ch:conclusion}

In this paper, we propose \savla{}, a symmetry-aware vision-language-action model that builds rotational symmetry into its architecture instead of learning it from demonstrations.
On top of a frozen pretrained vision-language backbone, we design an action head that is equivariant by construction, a typed interface that generates invariant and equivariant conditions, and a canonicalization head that brings the oblique view into a canonical frame.
Together these components form an end-to-end symmetry-aware VLA.
On the four LIBERO suites, \savla{} outperforms \groot{} by $5.1$ points on average, and the gap widens as the demonstration budget shrinks.
Under scene rotations on LIBERO-Goal, \savla{} improves the mean success rate from $41.5\%$ to $90.4\%$.

Several limitations remain.
Although the action head itself is equivariant to the full rotation group $\mathrm{SO}(3)$, the canonicalization head estimates only rotation about the gravity axis, and our rotation experiments cover only such rotations.
Moreover, the performance drop at large angles stems from the approximate canonicalization, whose parallax and field-of-view errors grow with the angle.

\section*{Impact Statement}
This paper presents work whose goal is to advance the field of machine learning and robot manipulation.
There are many potential societal consequences of our work, none of which we feel must be specifically highlighted here.

\bibliography{savla}
\bibliographystyle{IEEEtran}

\end{document}